\documentclass[lettersize,journal]{IEEEtran}
\usepackage{amsmath,amsfonts}
\usepackage{algorithmic}
\usepackage{algorithm}
\usepackage{array}
\usepackage[caption=false,font=normalsize,labelfont=sf,textfont=sf]{subfig}
\usepackage{textcomp}
\usepackage{stfloats}
\usepackage{url}
\usepackage{verbatim}
\usepackage{graphicx}
\usepackage{cite}
\usepackage{multirow}
\usepackage{booktabs}
\usepackage{pifont}
\usepackage{color}
\usepackage{diagbox}
\usepackage{bm}

\def\comment#1{{}}

\def\ie{{\em i.e.}}
\def\etc{{\em etc}}
\def\etal{{\em et al.}}

\begin{document}

\title{SDF-3DGAN: A 3D Object Generative Method Based on Implicit Signed Distance Function}

	
\author{Lutao Jiang$^*$,
	Ruyi Ji$^*$,
	Libo Zhang$^\dag$ \\
	\thanks{$*$The two authors contributed equally to this work \par
		$\dag$Corresponding author: Libo Zhang \par
		Lutao Jiang and Ruyi Ji is with the State Key Laboratory of Computer
		Science, Institute of Software Chinese Academy of Sciences, Beijing 100190,
		China, and also with the University of Chinese Academy of Sciences, Beijing
		101400, China (e-mail: lutao2021@iscas.ac.cn; ruyi2017@iscas.ac.cn).\par
		Libo Zhang is with the State Key Laboratory of Computer
		Science, Institute of Software Chinese Academy of Sciences, Beijing 100190,
		China (e-mail: libo@iscas.ac.cn).}
	
\thanks{This paper was produced by the IEEE Publication Technology Group. They are in Piscataway, NJ.}
\thanks{Manuscript received April 19, 2021; revised August 16, 2021.}}

\markboth{Journal of \LaTeX\ Class Files,~Vol.~14, No.~8, August~2021}%
{Shell \MakeLowercase{\textit{et al.}}: A Sample Article Using IEEEtran.cls for IEEE Journals}


\maketitle

\begin{abstract}
In this paper, we develop a new method, termed SDF-3DGAN, for 3D object generation and 3D-Aware image synthesis tasks, which introduce implicit Signed Distance Function (SDF) as the 3D object representation method in the generative field. We apply SDF for higher quality representation of 3D object in space and design a new SDF neural renderer, which has higher efficiency and higher accuracy. To train only on 2D images, we first generate the objects, which are represented by SDF, from Gaussian distribution. Then we render them to 2D images and use them to apply GAN training method together with 2D images in the dataset. In the new rendering method, we relieve all the potential of SDF mathematical property to alleviate computation pressure in the previous SDF neural renderer. In specific, our new SDF neural renderer can solve the problem of sampling ambiguity when the number of sampling point is not enough, \ie use the less points to finish higher quality sampling task in the rendering pipeline. And in this rendering pipeline, we can locate the surface easily. Therefore, we apply normal loss on it to control the smoothness of generated object surface, which can make our method enjoy the much higher generation quality. Quantitative and qualitative experiments conducted on public benchmarks demonstrate favorable performance against the state-of-the-art methods in 3D object generation task and 3D-Aware image synthesis task. Our codes will be released at https://github.com/lutao2021/SDF-3DGAN.
\end{abstract}


\section{Inroduction}

3D object generation has arguably emerged as one of the most intriguing research hotspots. This task is different from the reconstruction task. It requires the creation of an object that do not exist in the world from a random vector. With the advent of GAN \cite{goodfellow2014generative}, diverse follow-up architectures have popularized a variety of 2D image generation tasks \cite{karras2019style, karras2020analyzing, karras2021alias}. GAN endows neural networks with the capability of creation, \ie, generating the fake samples resembling real samples. However, its potential in 3D object generation task has yet to be further explored. As such, we observe that all existing methods for this task remain a extremely low level due to the limitations such as computation overhead, representation method, \etc.

\begin{figure}[!t]
	\centering
	\includegraphics[width=0.9\linewidth]{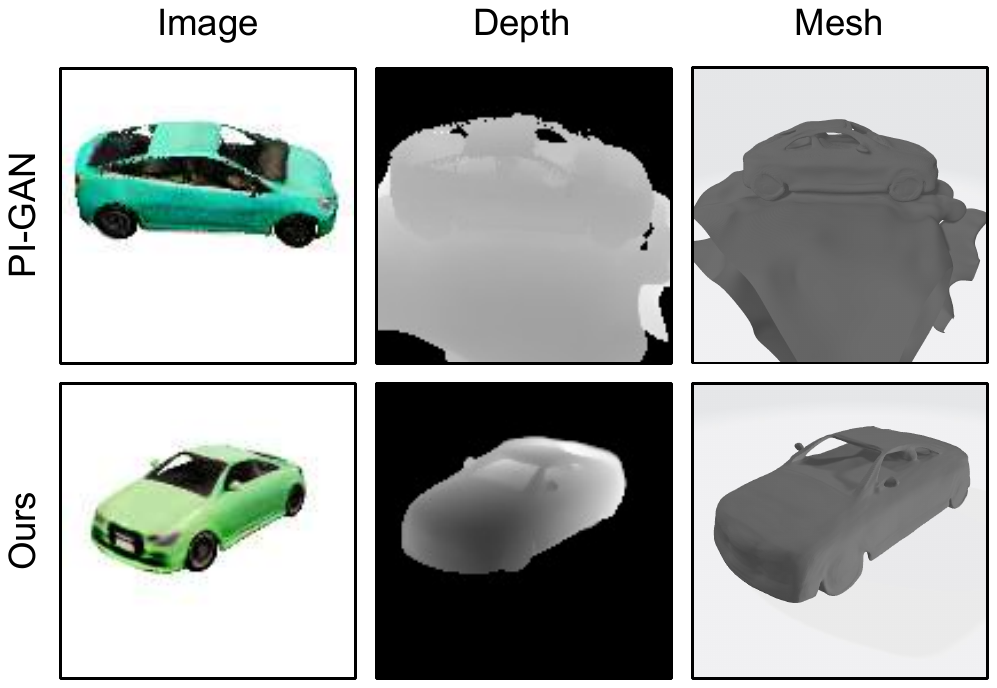}
	\caption{The comparison between PI-GAN (the 1$ ^{st} $ row) and ours (the 2$ ^{nd} $ row). The first column is RGB image. The second column is depth map. The third column is 3D mesh.}
	\label{fig:comparison}
\end{figure}

3D-GAN \cite{wu2016learning} is one representative of early methods. It generates a voxel model directly, but its spatial resolution is heavily restricted by the expensive memory consumption and low accuracy of voxel representation method. Recently, NeRF \cite{mildenhall2020nerf} has gained momentum in 3D representation domain. The main idea of NeRF is to utilize a simple Multi-Layer Perception (MLP) network to encode the implicit representation of a scene. To be specific, given the position coordinate and viewing direction of any point in space, MLP network predicts the volume density and RGB color of this point. After that, the color of ray can be accumulated by the volume rendering \cite{kajiya1984ray} algorithm. With the advance of NeRF, numerous NeRF-based generators have been proposed, such as GRAF \cite{schwarz2020graf}, PI-GAN \cite{chan2021pi} and GOF \cite{xu2021generative}. Theoretically, these methods enjoy unlimited spatial resolution and can generate information of arbitrary point. Nevertheless, the existing NeRF-based generators individually predict the volume density or opacity  of each point in space, neglecting the rich relationship between points. Such a schema easily incurs discontinuity defects on the real 3D model, and even some chaotic performance, as shown in Fig.~\ref{fig:comparison} (the 1$^{st}$ row).

Inspired by success of implicit Signed Distance Function (SDF)\cite{park2019deepsdf, jiang2020sdfdiff, atzmon2020sal, yariv2020multiview, wang2021neus}, we propose SDF-3DGAN, a method that introduce implicit Signed Distance Function (SDF) as the 3D object representation method for the 3D object generation and 3D-Aware image synthesis task. Different from the previous methods, which generates the voxel or the NeRF representation, we present to utilize SDF to represent the shape of an object. In general, SDF-3DGAN leverages an implicit SDF to represent an object and can render 2D RGB images from arbitrary angles. It is worth noting that the implicit SDF representation can be converted into other formats of 3D models, such as mesh. Compared to the voxel representation, the SDF representation can overcome its low representation accuracy due to its discrete nature in space. And the most important reason why our results are significantly better than previous NeRF-based methods (Fig.~\ref{fig:comparison}) is that the implicit SDF allows every point in space to perceive the presence of the surface and guarantees a stronger correlation between spatial points.

SDF is a function of the shortest distance from a point to the surface in space. The signed distance of a point outside the surface is defined as a positive value, while a point inside the surface is defined as a negative value. The zero-level set of a signed distance function is the surface of an object. In our design, we adopt a simple MLP network as the implicit signed distance function. Its input is the coordinate of a point and the output is signed shortest distance to the object surface. And we explicitly constrain its gradient's norm at every point to approach $1$, which turns this simple MLP network to be a implicit SDF. 

In the training stage, our overall training pipeline consists of two steps. Using the 2D car dataset as an example, we generate a 3D car model by giving the neural network a restricted random vector. To supervise its ability to convert arbitrary random vectors into different cars, we need to render them as 2D images at various angles by the rendering methods we have designed. The discriminator of GAN is then used to determine whether the generated image is similar to the 2D car images in the dataset. By achieving the goal of making the generated SDF look like a car at all angles, our generator can convert arbitrary random vectors into different 3D models of cars.

However, to render a 2D image with the resolution of $H \times W$, common neural volume rendering algorithms require to sample $H \times W$ rays, along with $N$ points on each ray. Considering a 2D image with $H=W=N=128$, we need to query the MLP network $2,097,152$ times to render such an image, resulting in the prohibitive calculation cost. To alleviate the computation burden, we design a new SDF rendering pipeline which takes full advantage of excellent mathematical properties of SDF to improve the sampling efficiency. And this rendering method is also can be applied in other SDF tasks.

Surprisingly, we notice concurrent method StyleSDF \cite{or2021stylesdf} shares the similar idea with us. Even though both leverage SDF as the representation method of an object, the focus of our two papers is significantly different. StyleSDF focuses on high-resolution ($1024 \times 1024$) face generation by training an overly complicated model similar to StyleGAN \cite{karras2019style}, which requires more computation resource to upsample the face resolution in the second stage, while its first stage is just amenable for low-resolution ($64 \times 64$) feature map generation. And two stages are optimized separately, which breaks the multi-view consistency in 3D-aware image synthesis. In contrast, we pay more attention to generate more realistic 3D objects and high-quality 3D-aware images simultaneously. Moreover, we propose to utilize the mathematical properties of SDF to reduce computation burden when rendering an image. And we can readily find the points and normal vectors of the surface. Therefore, we introduce normal loss to regularize the smoothness of surface, which allows our method for more realistic representation. In addition, with our improved rendering algorithms, which solves the problem of sampling error when the number of sampling points is too small (in the section IV.B.3), we can support the training of 360° datasets, not just forward-facing small-angle face datasets (StyleSDF only use these datasets). We validate the effectiveness of our method on a full angle dataset CARLA \cite{schwarz2020graf}, beyond the face domain. Moreover, our method produces higher quality and higher resolution feature maps than the first stage of StyleSDF. I believe that replacing its first stage generator with ours would yield better results if computational resources were available.

To sum up, the main contributions of this work can be summarized into five folds.
\begin{itemize}
	\item We present a novel method for 3D object generation, named SDF-3DGAN, which introduce implicit SDF as the representation method of 3D object in the generative field.
	\item We design a new SDF neural renderer for implicit SDF representation method, which can render the generated object to 2D RGB images more efficiently and accurately than previous SDF neural renderer. And this rendering method is also can be applied in other SDF tasks.
	\item The proposed rendering pipeline takes full advantage of mathematical properties of SDF to reduce computation cost and significantly improves the sampling efficiency. Besides, we avoid the dilemma that it is prone to locate the wrong surface position even cannot find surface when the sampling points are insufficient. 
	\item Thanks to the fact that we can readily find the desirable surface, we can constrain the variation of the surface normal vector, \ie, the surface SDF gradient, which greatly smooths the surface of the generated object.
	\item Quantitative and qualitative experiments conducted on the challenging benchmarks, including CARLA \cite{schwarz2020graf}, CelebA \cite{liu2015faceattributes}, BFM \cite{wu2020unsupervised}, show favorable performance against the state-of-the-art methods, validating effectiveness of our proposed pipeline and optimization algorithms.
\end{itemize}

The remainder of this paper is organized as follows. Section \ref{sec:related work} briefly reviews significant works relevant to our method. In Section \ref{sec:preliminaries}, we have briefly introduced some of the key concepts. In Section \ref{sec:method}, we describe the proposed pipeline and method in detail. In Section \ref{sec:experiments}, the extensive experimental results are reported and analyzed, including the comparison between the proposed method and existing cutting-edge methods, validation of generation ability, and the comprehensive analysis of the ablation study. Finally, Section \ref{sec:conclusion} draws the conclusions of the proposed method and summarizes several potential directions for the future research.

\section{Related Work}\label{sec:related work}
In this section, we briefly review three major research directions closely related to our method, \ie, 3D generative model, implicit signed distance function, and generative adversarial networks.

\subsection{3D Generative Model}
As one of seminal methods, 3D-GAN \cite{wu2016learning} proposes to use GAN for the task of 3D object generation, where it relies on the generator to produce a voxel of a 3D object. VON \cite{zhu2018visual} proposes to disentangle object representation. Concretely, it first generates a 3D model by 3D-GAN, then computes a perspective-specific 2.5D sketch by differentiable projection, and finally generates a 2D image by texture network.
However, both 3D-GAN and VON require 3D supervision, and the memory-sensitive drawback dramatically limits their spatial resolution. HoloGAN \cite{nguyen2019hologan} proposes a method only requiring 2D unlabeled images to generate 3D-aware images. Unfortunately, it cannot explicitly generate a 3D model in space. The arrival of NeRF \cite{mildenhall2020nerf} has revolutionized the paradigm of retrieving the volume density and RGB color of any point in space via a neural network. For example, GRAF \cite{schwarz2020graf} is the first method to introduce NeRF into the 3D object generation task. It leverages a NeRF-like MLP structure to generate 3D object. PI-GAN \cite{chan2021pi} designs a mapping network similar to StyleGAN \cite{karras2019style}, and introduces FiLM \cite{perez2018film} and SIREN \cite{sitzmann2020implicit} activation functions. The follow-up method GOF \cite{xu2021generative} changes the volume density of PI-GAN to occupancy, and proposes an algorithm to allow model aware of the surface of the object. ShadeGAN \cite{pan2021shading} attaches the illumination model to PI-GAN and designs a surface tracking CNN-based network to improve sampling efficiency. Likewise, there exist some two-stage methods \cite{gu2021stylenerf, zhou2021cips, or2021stylesdf} that utilize volume rendering algorithms to produce low-resolution features which are then upsampled to a high resolution image. It is worthy to note that all these implicit representation methods can be converted to other 3D representation methods such like mesh.

\subsection{Implicit Signed Distance Function}
With the development of implicit neural representations (INR), ever-increasing research attention is shifted to adopt deep neural networks as implicit signed distance functions. Considering that a deep neural network can serve as a continuous function, DeepSDF \cite{park2019deepsdf} pioneers to introduce deep neural network into 3D model representation and reconstruction. To get free from 3D supervision in single-view or multi-view 3D reconstruction task, SDFDiff \cite{jiang2020sdfdiff} proposes to use a differentiable method to render 3D shapes represented by signed distance functions. SAL \cite{atzmon2020sal} designs a novel method based on implicit shape representation for surface reconstruction tasks from an un-oriented point cloud. IDR \cite{yariv2020multiview} introduces a neural network architecture to disentangle geometry and appearance, allowing model to encode more complex lighting conditions and materials. NeuS \cite{wang2021neus} proposes a new SDF-based volume rendering method, which maps signed distance to its proposed s-density and generalize to the volume rendering algorithm. 

\subsection{Generative Adversarial Networks}
Ian GoodFellow \etal \cite{goodfellow2014generative} propose generative adversarial networks, which consist of a generator and a discriminator.
Given a random noise, the generator network can convert it to fake data which tries to resemble real data as far as possible. The discriminator network is tasked with distinguishing whether the input data comes from real data or fake data. The two parts evolve together in an adversarial manner, ending up with a generator network capable to produce high quality data. GAN has been applied to a variety of tasks, particularly in the field of 2D images, such as image generation \cite{karras2017progressive, karras2019style, karras2020analyzing, karras2021alias, skorokhodov2021adversarial, zhang2021lightweight, ma2021spatial, chen2021pman}, inpainting \cite{demir2018patch, xu2020e2i}, dehazing \cite{zhao2020pyramid, wang2021tms}, \etc. Specifically, progressive GAN \cite{karras2017progressive} proposes a gradual transition training scheme from low-resolution to high-resolution. Following a coarse-to-fine paradigm, the training process starts by learning coarse-grained features and then pays more attention to the fine-grained features. The StyleGAN series \cite{karras2019style, karras2020analyzing, karras2021alias} not only significantly improve the quality of generated images, but also spark significant improvements in image reconstruction and editability. INR-GAN \cite{skorokhodov2021adversarial} proposes to leverage a continuous function which can generate an image by querying the coordinates of each pixel.

\section{Preliminaries}\label{sec:preliminaries}
To better understand the proposed method, we first briefly introduce a few important concepts, \ie, the representative neural rendering method, \textit{NeRF}, the conventional sampling strategy on the ray, \textit{hierarchical sampling}, and Signed Distance Function (SDF). If you are already familiar with these concepts, we recommend skipping this section.

\begin{figure}[t]
	\centering
	\includegraphics[width=2in]{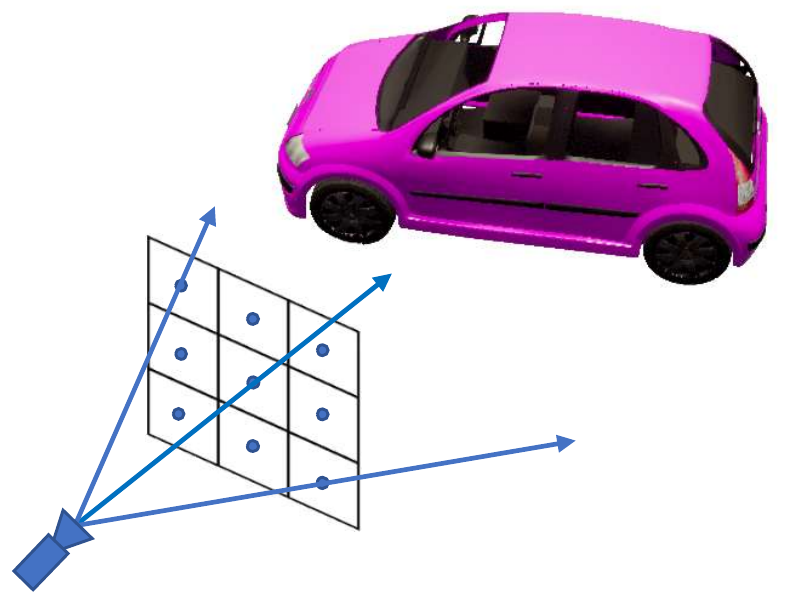}
	\caption{The illustration of ray sample. The grid in front of the camera is the imaginary imaging plane. And each grid represents one pixel. We need to emit some rays from camera source and each ray represents a pixel.}
	\label{fig:ray_sample}
\end{figure}

\subsection{NeRF (Neural Radiance Field)}
NeRF readily represents a 3D object or a scene through a simple neural network. It predicts the volume density and RGB color values for each point in a scene, then the volume rendering algorithm is used to render the observed image at the given viewpoint. Given the coordinate $\bm{x}$ of a point and the observation direction $\bm{d}(\theta,\phi)$ ($\theta$ and $\phi$ are the pitch and yaw respectively), NeRF queries the volume density $\sigma$ and the RGB color $\bm{c}$ of a point through MLP. To render an RGB image, we first set view-dependent rays, then sample points along these rays and query their volume density $\sigma$ and RGB color $\bm{c}$ respectively. Finally, the classic volume rendering technique \cite{kajiya1984ray} is utilized to calculate the color of each ray. Each ray corresponds to a pixel in the 2D image, as shown in Fig. \ref{fig:ray_sample}. The specific calculation of each ray is mathematically reformulated as follows:
\begin{equation}
	\begin{split}
		\hat{\bm{C}}(\bm{r})=\sum_{i=1}^{N} T_{i}\left(1-\exp \left(-\sigma\left(\bm{x}_{i}\right) \delta_{i}\right)\right) \bm{c}\left(\bm{x}_{i}, \bm{d}\right), \\
		\text { where } T_{i}=\exp \left(-\sum_{j=1}^{i-1} \sigma\left(\bm{x}_{j}\right) \delta_{j}\right) 
		\label{eq:color_calculate}
	\end{split}
\end{equation}
where $\sigma(\bm{x}_{i})$ stands for the volume density of the point $\bm{x}_{i}$, $\bm{c}\left(\bm{x}_{i}, \bm{d}\right)$ is the RGB color of point $\bm{x}_{i}$ conditioned on observing direction $\bm{d}$, $\delta_{i}$ is the distance between the two points of $\bm{x}_{i}$ and $\bm{x}_{i+1}$, and $\bm{r}$ is the ray which is represented $\bm{r}(t)=\bm{o}+t\bm{d}$, $\bm{x}_{i}=\bm{r}(t_i)$.

\subsection{Hierarchical Sample}
To calculate the color of each ray via the classical rendering algorithms, there is a requirement to sample points along the ray. NeRF offers a hierarchical sampling strategy. That is, in the first round, the sampling points along a ray are drawn from a uniform distribution. After that, we calculate the weight of $i$-th sampling point through the formula $T_{i}\left(1-\exp \left(-\sigma\left(\bm{x}_{i}\right) \delta_{i}\right)\right)$ on each ray. Formulating above-acquired weights as the probability density, the second round sampling is performed along this ray. Finally, all points sampled in the above two rounds contribute to the calculation of color belonging to this ray in Equ.~(\ref{eq:color_calculate}).

\subsection{SDF (Signed Distance Function)}
SDF refers to a function of the shortest distance from a point to the surface in space. Following the common practice, the signed distance of points outside the surface are positive and negative for those inside the surface. The zero-level set of a signed distance function is the surface of an object. Theoretically, if a function is with gradient's norm $1$ at every point in space, then it can serve as the SDF of an object. Likewise, if we explicitly constrain the gradient's norm of network's output to be $ 1 $ with respect to the input points, such a neural network turns to be an implicit SDF. Obviously, the gradient of SDF of a point on the surface is its normal vector. Taking a ball of radius $1$ whose center is located at the origin of coordinate system as an example, its SDF can be mathematically formulated as $f(x,y,z) = \sqrt{x^2+y^2+z^2}-1$. Its surface is the zero-level set of SDF, which is written as $\{(x,y,z)|f(x,y,z) = \sqrt{x^2+y^2+z^2}-1=0\}$.

\section{Method}\label{sec:method}

\begin{figure*}[t]
	\centering
	\includegraphics[width=7in]{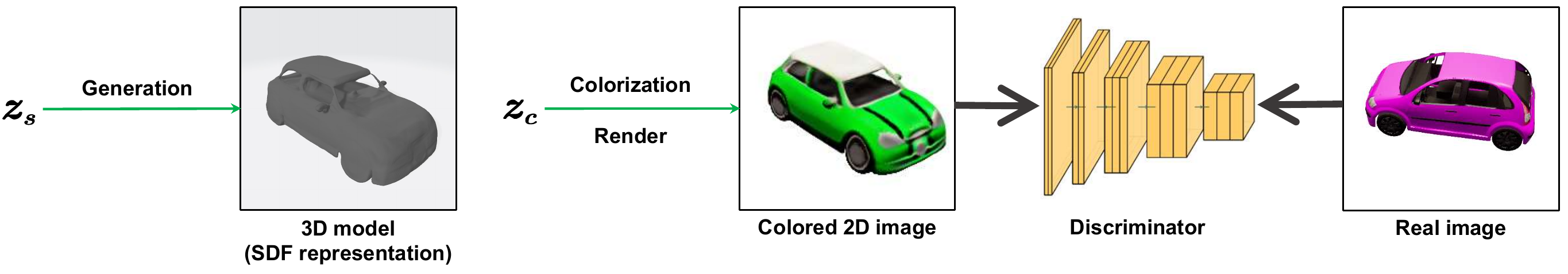}
	\caption{The illustration of whole training pipeline. We first generate a 3D model. Then we render 2D RGB images at all angles to train. Finally, we use the generated 2D image and the real image of dataset to apply the GAN training method.}
	\label{fig:full_pipeline}
\end{figure*}

As in Fig.~\ref{fig:full_pipeline}, our overall training process is divided into two stages. First, given shape code $\bm{z_s}$, which is a  random vector, we can generate an object represented by $SDF(\bm{x};\bm{z_s})$. And this SDF is a conditioned network. Second, given color code $\bm{z_c}$, which is also a random vector, we can determine the surface color of this object. Then, we can render this object to a 2D image at any angle by our new designed rendering algorithm. Through training, we leave the discriminator without the ability to distinguish the difference between the generated arbitrary-angle images and the images in the dataset, \ie the generated images are close enough to the images of the given dataset. The object generated in space thus satisfies the shape we want to generate. Considering there are large number of symbols in the next paper, we have therefore summarised a symbol table in Table \ref{tab:symbol table}.

\begin{table}
	\scriptsize
	\caption{symbol table}
	\label{tab:symbol table}
	\begin{tabular}{c|l}
		\hline
		          Symbol            & \multicolumn{1}{c}{Illustration}                                     \\ \hline
		        $\bm{z_s}$          & Shape code. For controlling the shape of randomly generated objects. \\
		        $\bm{z_c}$          & Color code. For controlling the surface color of generated objects.  \\
		         $\bm{x}$           & Three-dimensional coordinates.                                       \\
		         $\bm{d}$           & The observation direction of a point or the observation of a ray.    \\
		$\alpha$ or $\alpha(\cdot)$ & The opacity or the opacity of a point $\bm{x}$.                      \\
		$\bm{c}$ or $\bm{c}(\cdot)$ & The color of a point $\bm{x}$.                                       \\
		            $s$             & The signed distance of a point $\bm{x}$ from a surface.              \\
		         $\bm{o}$           & The ray origin of a ray.                                             \\
		         $\delta$           & The sampling interval.                                               \\ \hline
	\end{tabular} 
\end{table}
 
\subsection{Generate Object}

\begin{figure*}[!h]
	\centering
	\includegraphics[width=7in]{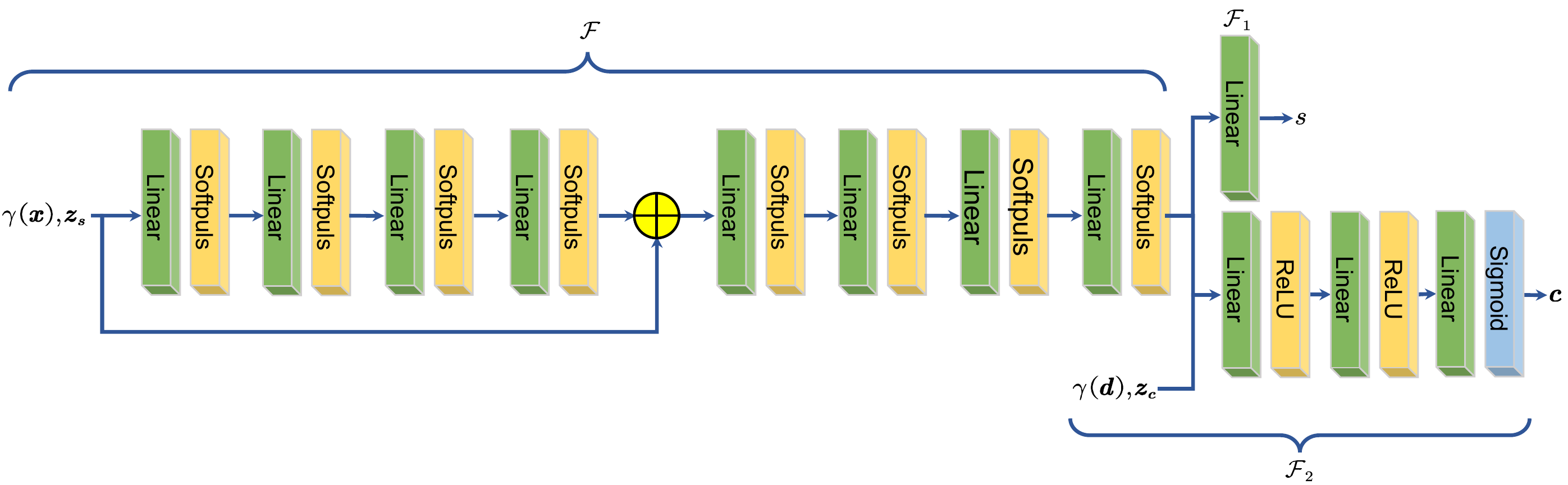}
	\caption{The MLP network structure. $\gamma(\cdot)$ is the position encoding function as same as NeRF. The Green linear block represents linear layer. The yellow block means activation functions, including softplus and ReLU. The blue block is the sigmoid function, which aims to make the output in [0,1].}
	\label{fig:model}
\end{figure*}

We first generate an object in space, which is represented by the SDF neural network. As shown in Fig.~\ref{fig:model}, we use a simple MLP network as the SDF. This neural network can be expressed as the following equation
\begin{equation}
	\begin{aligned}
		\bm{V}&=\mathcal{F}(\bm{x};\bm{z_s}) \\
		s&=\mathcal{F}_1(\bm{V}) \\
		\bm{c}&=\mathcal{F}_2(\bm{V},\bm{d};\bm{z_c})
	\end{aligned}
	\nonumber 
\end{equation}

where $\mathcal{F}$, $\mathcal{F}_1$ and $\mathcal{F}_2$ are some linear layers in the Fig.~\ref{fig:model}, $\bm{x}{\in}\mathbb{R}^{3}$ is the 3D space coordinate, $\bm{z_s}{\in}\mathbb{R}^{128}$ is the shape code, $\bm{V}{\in}\mathbb{R}^{256}$ is the intermediate feature vector, $s$ is the signed distance, $\bm{d}{\in}\mathbb{R}^{3}$ is the observation direction of the point, $\bm{z_c}{\in}\mathbb{R}^{128}$ is the color code and $\bm{c}{\in}\mathbb{R}^{3}$ is the RGB color of point. And in the training stage, both $\bm{z_s}$ and $\bm{z_c}$ obey standard Gaussian distribution.

Through fixing the input $\bm{z_s}$ to the neural network $\mathcal{F}_1(\mathcal{F}(\cdot))$, this neural network can be seen as the function $SDF(\bm{x};\bm{z_s})$, where $\bm{x}$ is the argument of this function. Different $\bm{z_s}$ represent different objects. Taking the car dataset as an example, after training is completed, changing different $\bm{z_s}$ will change the shape of the car. For training, we need to render the object to 2D RGB images at arbitrary angle. Therefore, we attach a color branch to the network to achieve this goal. We input the position feature vector, observation direction and color code to this branch to obtain the RGB value of the point. Similarly, different $\bm{z_c}$ can change the color of the object.

\subsection{Render the Generated Object to 2D RGB Image}

\subsubsection{\textbf{The Proposed SDF Neural Renderer}}
\begin{figure*}[!t]
	\centering
	\includegraphics[width=7in]{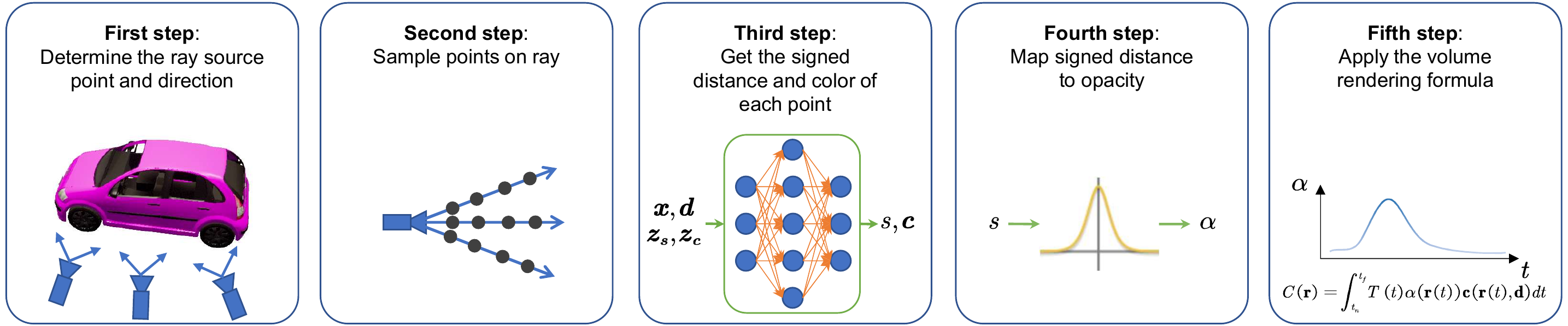}
	\caption{Our rendering pipeline. First, the camera need to be placed and aimed at the object to emit some rays, where each ray is a pixel. Second, in order to calculate the color of each ray, it is necessary to integrate the RGB colors of the rays according to their opacity $\alpha$ at different locations, and to improve the efficiency of the integration it is necessary to sample some points on the ray. Third, given the shape code $\bm{z_s}$ and appearance code $\bm{z_c}$, the coordinates $\bm{x}$ of the points sampled earlier and the direction $\bm{d}$ of the ray are input to the network to obtain the signed distance $s$ and the RGB color $\bm{c}$. Fourth, we need to map the signed distance $s$ to opacity $\alpha$. Fifth, the volume rendering formula is applied to use approximate integration to accumulate the colors on the ray to obtain the final color. Getting colors of all rays, we finish to render the 2D image.}
	\label{fig:pipeline}
\end{figure*}

The overall flow-chart of the proposed SDF neural renderer is illustrated in Fig.~\ref{fig:pipeline}. To render an image, we first sample the camera position in space. For convenience, we sample the camera position on a sphere of radius $1$ which is located at the origin of the coordinate system. Similar to PI-GAN \cite{chan2021pi}, we then turn the camera direction towards the origin of coordinates and emit $H \times W$ rays (each ray corresponds to one pixel). After that, we sample points along the ray for subsequent integration calculation. Next, the position coordinate together with observation direction are fed into the well-designed SDF neural network to compute the according signed distance and RGB color. To adapt to the classical volume rendering formula, we project the signed distance to opacity. Finally, we leverage the volume rendering formula to calculate the color of each ray, which is equivalent to rendering the entire picture. Next, we will introduce the color calculation method and point sampling method in detail.

\subsubsection{\textbf{Map Signed Distance to Opacity and Calculate Color}}
As discussed above, to adapt to the classical volume rendering formula, we explicitly map signed distance to opacity $\alpha\in(0,1]$ and calculate the color of each ray via Equ.~(\ref{eq:color_calculate2}).
\begin{equation}
	\hat{\bm{C}}(\bm{r}; \bm{z_s}, \bm{z_c})=\sum_{i=1}^{N} \alpha(\bm{x}_{i}; \bm{z_s}) \prod_{j<i}(1-\alpha(\bm{x}_{j}; \bm{z_s})) \bm{c}\left(\bm{x}_{i}, \bm{d}; \bm{z_s}, \bm{z_c}\right)
	\label{eq:color_calculate2}
\end{equation}
where $\bm{r}$ is a ray, and $N$ $\bm{x_i}$ are the points on the ray, which are arranged from near to far from the position of the ray origin. $\alpha(\bm{x_i};\bm{z_s})$ means the opacity of point $\bm{x_i}$ conditioned on shape code $\bm{z_s}$, $\bm{c}(\bm{x_i}, \bm{d};\bm{z_s},\bm{z_c})$ is the color of point $\bm{x_i}$ with observation direction $\bm{d}$ conditioned on color code $\bm{z_c}$.

Intuitively, the opacity of a point should be larger when it approaches to the surface of an object. When far from the surface, its opacity should approach to $0$. To satisfy the above characteristics, we design a mapping function $\mathcal{M}(\cdot)$, which is tasked with converting signed distance of a point to its opacity. 
Formally, a desirable mapping function should be with following three characteristics: 
\begin{enumerate}
	\item $\mathcal{M}(s)=\mathcal{M}(-s)$
	\item Monotonically increasing at $(-\infty,0)$ and monotonically decreasing at $(0,+\infty)$
	\item Approach to $0$ when the signed distance is far from $0$ and equal to $1$ when the signed distance is $ 0 $.
\end{enumerate}

\begin{figure}[t]
	\centering
	\includegraphics[width=\linewidth]{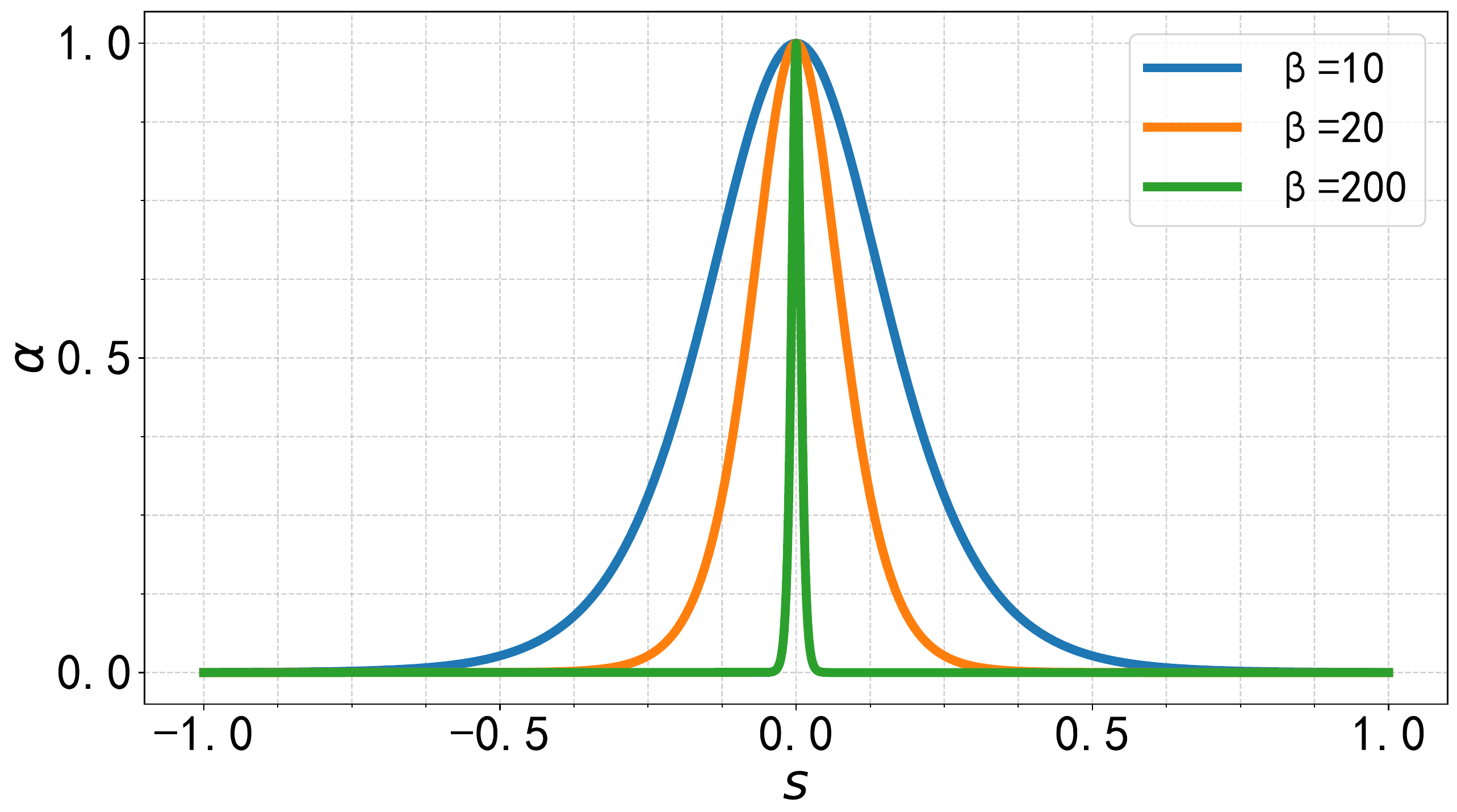}
	\caption{The illustration of $\mathcal{M}(s;\beta)$. The horizontal axis indicates the signed distance $s$. The vertical axis represents the opacity $\alpha$. The larger $\beta$ is, the more compact the curve is.}
	\label{fig:map_function_curve}
\end{figure}

Empirically, we observe that the derivative function of sigmoid satisfies the above conditions well. Since such a function varies in the range of $(0,0.25]$, we regularize it by a scale factor of four. Thereby, the mapping function can be given as Equ.~(\ref{eq:map_function}), where $\beta$ is a learnable parameter increasing with the progress of training, which ensures the function $\mathcal{M}(s)$ to shrink towards $s=0$ , as shown in Fig. \ref{fig:map_function_curve}.

\begin{equation}
	\mathcal{M}(\mathrm{s} ; \beta)=4 \cdot \operatorname{sigmoid}(\beta \cdot \mathrm{s}) \cdot (1-\operatorname{sigmoid}(\beta \cdot \mathrm{s}))
	\label{eq:map_function}
\end{equation}

\subsubsection{\textbf{Sampling Method and Accurate Sampling Strategy}}
Assuming a ray $\bm{r}(t)=\bm{o}+t\bm{d}$, where $\bm{o}$ refers to source point, $\bm{d}$ represents the ray direction and $t$ is a ray depth scalar, we sample points along the ray to proceed volume rendering via Equ.~(\ref{eq:color_calculate2}). More specifically, we need to sample $N$ points, \ie~ray depth scalar $t$, within an interval for subsequent calculation on the ray. Upon the shortest distance to surface acquired by SDF, we leverage ray marching algorithm to search the approximate position of the surface. Mathematically, we apply the Equ.~(\ref{eq:ray_marching}) to iteratively search the surface location.
\begin{equation}
	t_{i+1} = t_{i} + SDF(\bm{o}+t_i\bm{d})
	\label{eq:ray_marching}
\end{equation}

\begin{figure}[!h]
	\centering
	\includegraphics[width=\linewidth]{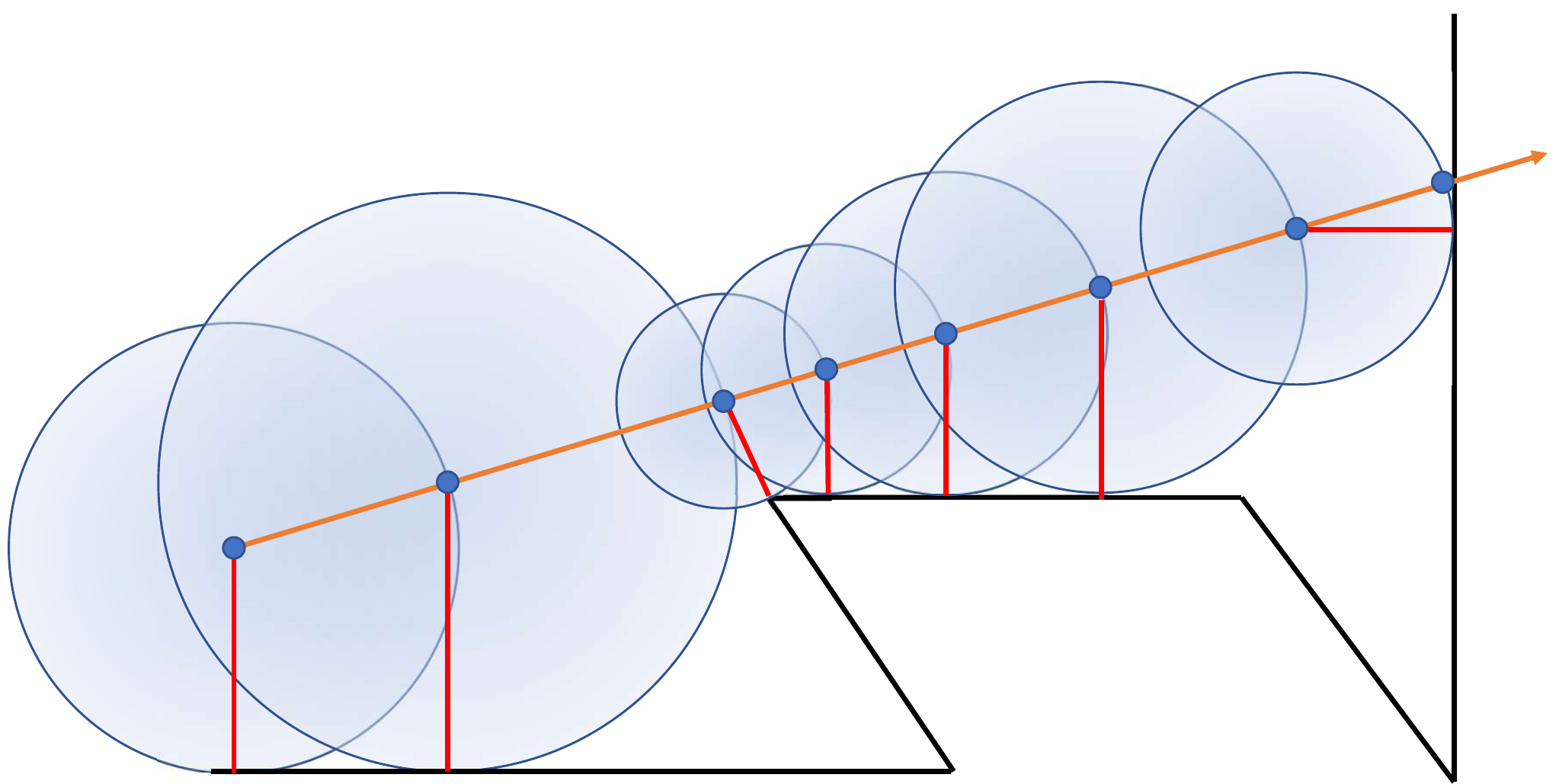}
	\caption{The illustration of ray marching. The black line  in the diagram is the surface of an object. The orange line is a ray emitted from the camera's point of view. Each circle in the diagram is with the shortest distance to surface as its radius, and the red line is the shortest distance. We start with a starting depth $t_0$ on the ray (the blue dot on the left-most side), and query the signed distance value (the length of red line) of this point through MLP. Then add this value to $t_0$ to get $t_1$ (the second dot from left). Because the signed distance is the shortest distance from the point to the surface, moving forward this distance on the ray will not cross the surface. After some iterations of this process, we can find the approximate surface position (the right-most dot).}
	\label{fig:ray_marching}
\end{figure}

Obviously, for a point, marching forward with the shortest distance to surface on the ray will never cross the surface. It reaches the surface only if the direction of the ray coincides with that of the shortest distance. Fig. \ref{fig:ray_marching} presents a typical example for better understanding. After numerous rounds of computation, we figure out the approximate ray depth $t_d$ is within the acceptable error range. And we set this depth value $t_d$ to be the center of sampling interval. Then, we uniformly sample points in the $[t_d-\delta,t_d+\delta]$. $\delta$ refers to a hyper-parameter, which decreases as training proceeds. 

\begin{figure}[!h]
	\centering
	\includegraphics[width=\linewidth]{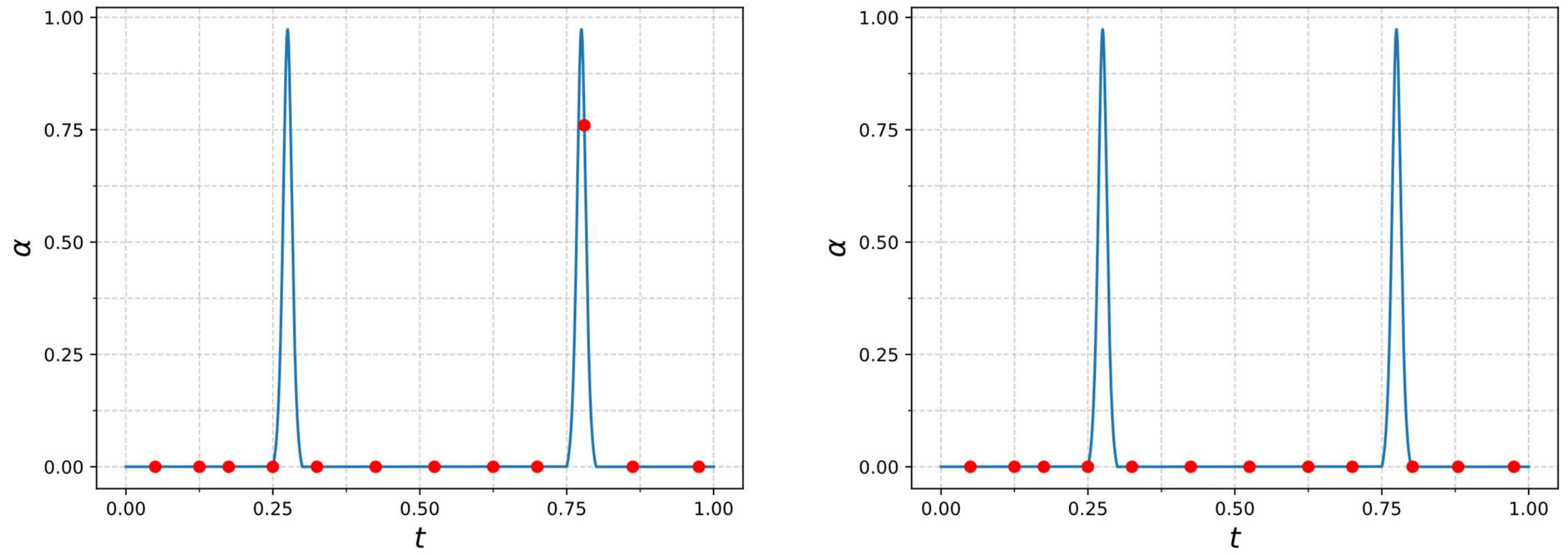}
	\caption{Two cases of sampling error. The horizontal axis indicates the ray depth scalar $t$ in $\bm{r}(t)=\bm{o}+t\bm{d}$. The vertical axis represents the opacity $\alpha$. The left one find the error place of the second surface (one ray may pass through two front and back surfaces). The right one doesn't find a place where the opacity is large. But we need to sample on the first peak.}
	\label{fig:two_error}
\end{figure}

As $\beta$ in Equ.~(\ref{eq:map_function}) increases in the training process, the $\alpha$ curve near the surface gets narrower gradually (refer to Fig.~\ref{fig:map_function_curve}). As a result, only points very close to the surface have an opacity far from $ 0 $. Therefore, when the sampling points are insufficient in the first round, it would be troublesome to find a correct surface or cannot find any surface, as illustrated in Fig.~\ref{fig:two_error}. As a result, when we calculate colors through Equ.~(\ref{eq:color_calculate2}), there exists two possible problems, (1) all points tend to be with a weight of $ 0 $, (2) points close to the second surface would have larger weights. As a result, the generated image would be with discontinuous points, which in turn poses challenges for the stable training (please refer to the first two rows of Fig.~\ref{fig:ablation_accuracy_sampling}).

\begin{figure}[t]
	\centering
	\includegraphics[width=\linewidth]{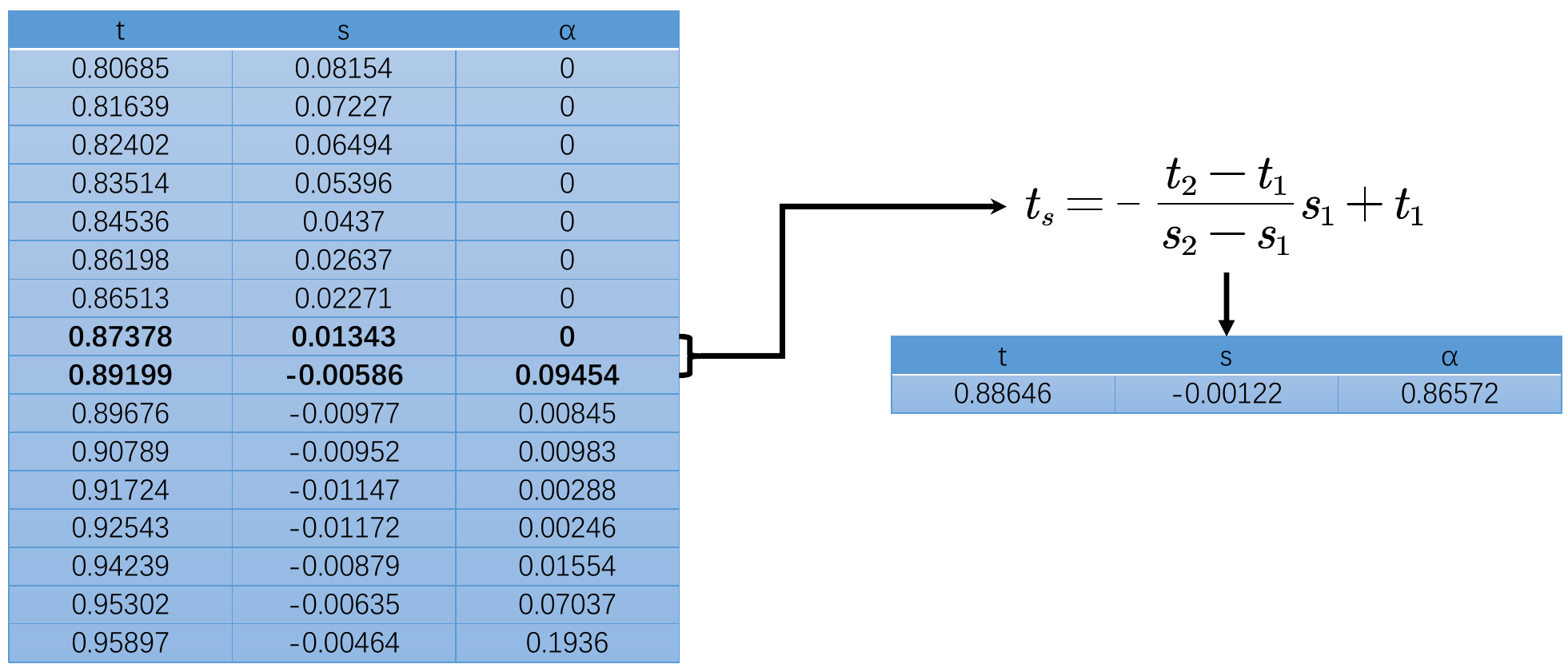}
	\caption{An example of accurate sampling. These values come from the actual experiment. According to the algorithm we designed, the two bold lines are the selected ($t_1, s_1$) and ($t_2, s_2$). And after our algorithm, we find the surface position more precisely.}
	\label{fig:accurate_sampling_example}
\end{figure}

To solve the above issues, we propose an accurate sampling strategy to ensure the sampling efficiency and accuracy. Concretely, on a ray $\bm{r}(t)=\bm{o}+t\bm{d}$, we first locate the two points which are the first pair of signed distance values varied from positive to negative in the previous uniform sampling procedure after ray marching. We denote such two points as $(t_1,s_1)$, $(t_2,s_2)$ respectively, where $s_1$ and $s_2$ indicate the signed distance values at positions $t_1$ and $t_2$, $s_1>0$, $s_2<0$. We project these two points in the coordinate system to compute the primary root of the line through Equ.~(\ref{eq:root}).
\begin{equation}
	t_s=-\frac{t_2-t_1}{s_2-s_1} s_1 + t_1
	\label{eq:root}
\end{equation}
where $t_s$ is the accurate surface that we locate. We provide an example for better understanding in Fig.~\ref{fig:accurate_sampling_example}. Empirically, such simple strategy not only solves the above two issues, but also serves as an alternative high-efficiency choice in hierarchical sampling (refer to section \ref{sec:preliminaries}-B). The ablation study in section \ref{section:Accurate sampling} demonstrates the superiority over its existing counterparts.

\subsection{Training Strategy}
When we render the object to 2D images, we can change the number of sampling rays, \ie, the number of pixels. One ray can be seen as one pixel. Therefore, we can render images at any resolutions theoretically. But more pixels means that we need to sample more rays and points to calculate the RGB values of each ray. Restricted by hardware, we only render the 2D images at $32 \times 32$ to $128 \times 128$ resolution.

Enjoying ability to render an image at any resolution, we borrow an off-the-shelf progressive discriminator used by PI-GAN \cite{chan2021pi} to first learn the coarse-grained content at a low resolution, and then gradually increase the resolution to learn more details. The training procedure is divided into three stages, which consist of $32 \times 32$, $64 \times 64$ and $128 \times 128$ resolutions. And as the rendered image resolution gradually increases, we shrink the range of sampling interval and reduce the number of sampling points to relieve the computation cost.

\subsection{Loss Function}

\paragraph{GAN Loss}
Let the generator be $G_{\theta_{1}}(\bm{z_s},\bm{z_c}, \xi)$, where $\theta_{1}$ is the learnable parameter and $\xi$ is the camera pose. As such, the discriminator is $D_{\theta_{2}}(\cdot)$, where $\theta_{2}$ refers to the learnable parameter.
The input of discriminator is either a real image $\mathcal{I}$ from dataset or a generated image from $G_{\theta_{1}}(\bm{z_s},\bm{z_c}, \xi)$. We adopt non-saturating GAN loss with R1 regularization \cite{mescheder2018training} as follows:
\begin{equation}
	\begin{split}
		\mathcal{L}_{GAN}=\mathbb{E}_{\bm{z_s} \sim p_{z_s}, \bm{z_c} \sim p_{z_c}, \xi \sim p_{\xi}}\left[f\left(D_{\theta_{2}}\left(G_{\theta_{1}}(\bm{z_s},\bm{z_c}, \xi)\right)\right)\right] \\
		+\mathbb{E}_{I \sim p_{\mathcal{D}}}\left[f\left(-D_{\theta_{2}}(\mathcal{I})\right)+\lambda\left|\nabla D_{\theta_{2}}(\mathcal{I})\right|^{2}\right] \\
		\text { where } \quad f(u)=-\log (1+\exp (-u))
	\end{split}
	\label{eq:GAN_loss}
\end{equation}

\paragraph{Eikonal Loss}
The Eikonal Loss in Equ.~(\ref{eq:eikonal}) is utilized to ensure network learns a meaningful SDF. In another word, we make efforts to ensure that the norm of the gradient vector at each point in space converge to $ 1 $.

\begin{equation}
	\mathcal{L}_{Eikonal}=\mathbb{E}_{\bm{x}}\left(\left\|\nabla_{\bm{x}} f(\bm{x} ; \bm{z_s})\right\|-1\right)^{2}
	\label{eq:eikonal}
\end{equation}

\paragraph{Normal Loss}
Further, we leverage the $ \ell_2 $ Normal Loss in Equ.~(\ref{eq:normal}) to smooth generated 3D model. 
\begin{equation}
	\mathcal{L}_{Normal}=\frac{1}{N}\sum_{\bm{x_{s}} \in \mathcal{S}}\left\|\nabla_{\bm{x_s}}\left(\bm{x_{s}}, \bm{z_s}\right)-\nabla_{\bm{x_s}}\left(\bm{x_{s}}+\bm{\epsilon}, \bm{z_s}\right)\right\|
	\label{eq:normal}
\end{equation}
where $N$ is the number of points involved in the calculation, $\mathcal{S}$ denotes the set of points on the surface and $\epsilon$ indicates the micro-disturbances. Concretely, the $ \ell_2 $ normal loss constraint is imposed on the normal vector variation of the surface. Exactly, the gradient of SDF at a point is the normal vector of this point.

In summary, the final loss function is:
\begin{equation}
	\mathcal{L}=\mathcal{L}_{GAN} + \lambda_{Eikonal}\mathcal{L}_{Eikonal} + \lambda_{Normal}\mathcal{L}_{Normal}
\end{equation}
where $\lambda_{Eikonal}$ and $\lambda_{Normal}$ are two hyper-parameters to balance the above loss items.

\section{Experiments}\label{sec:experiments}
In this section, we conduct extensive experiments to verify the effectiveness of our proposed method. Specifically, we first describe the three challenging datasets involved in this paper. Then, we introduce the implementation details of our experiments. Next, the comparisons between the proposed method and existing works are reported and analyzed qualitatively. Finally, comprehensive ablation studies are performed to help understand the proposed method better.

\subsection{Datasets}
To evaluate the effectiveness and generation ability of our method and for fair comparison to other methods, we conduct extensive experiments on three widely used and challenging datasets, including CARLA \cite{schwarz2020graf}, CelebA \cite{liu2015faceattributes}, BFM \cite{wu2020unsupervised}. These three datasets are used widely by most SOTA methods. The CARLA dataset is derived from GRAF \cite{schwarz2020graf} where the Carla Driving simulator \cite{dosovitskiy2017carla} is used to render $ 10,000 $ images of $ 18 $ car models and each image has a diverse appearance. The cars are all in the center of image and the background is white. The camera position of the rendered image is sampled on an upper hemisphere uniformly, which ranges from 0°-360° and 0°-85° for azimuth and polar angle, respectively. CelebA contains over $ 200,000 $ real-world images of faces without camera parameters. It contains images of face at various forward angles, but does not contain any pitch and yaw angle information. Following the common practice in prior works, we only use the head part patches cropped from original images in our experiments. With regard to BFM dataset, it is a collection of $ 200,000 $ face images generated by unsup3d \cite{wu2020unsupervised} using the Basel Face Model \cite{paysan20093d}. Different form CelebA, it is a simulation dataset, therefore, it lacks a lot of information on facial detail. Notably, during the training, we only use the original RGB color information and do not rely on any extra information of the above datasets.

\subsection{Implementation Details}
\begin{table*}[]
	\centering
	\caption{Experimental settings for CARLA, CelebA and BFM datasets respectively.}
	\label{tab:training details}
	\resizebox{\linewidth}{!}{
	\begin{tabular}{c|cccccccccc}
		\hline
		\multicolumn{1}{c|}{Datasets} & \multicolumn{1}{c}{Iteration} & \multicolumn{1}{c}{Resolution} & \multicolumn{1}{c}{Batch Size} & \multicolumn{1}{c}{Sampling Method} & \multicolumn{1}{c}{$\delta$} & \multicolumn{1}{c}{Point} & \multicolumn{1}{l}{Lr\_G} & \multicolumn{1}{c}{Lr\_D} & \multicolumn{1}{c}{$\lambda_{Eikonal}$} & \multicolumn{1}{c}{$\lambda_{Normal}$} \\ \hline
		   \multirow{3}{*}{CARLA}     &             0-20k             &               32               &               32               &             Coarse+Fine             &             0.3              &           32+32           &           4e-4            &           4e-4            &          \multirow{3}{*}{0.5}           &                   0                    \\
		                              &            20k-60k            &               64               &               24               &           Coarse+Accurate           &             0.15              &           32+1            &           4e-4            &           4e-4            &                                         &                  1.0                   \\
		                              &            60k-90k            &              128               &               16               &           Coarse+Accurate           &             0.075             &           16+1            &           4e-4            &           4e-4            &                                         &                  1.0                   \\ \hline
		   \multirow{3}{*}{CelebA}    &             0-50k             &               32               &              112               &             Coarse+Fine             &             0.12             &           32+32           &           4e-4            &           4e-4            &          \multirow{3}{*}{0.5}           &                   0                    \\
		                              &            50k-90k            &               64               &              112               &           Coarse+Accurate           &             0.06             &           16+1            &           4e-4            &           2e-4            &                                         &                  0.02                  \\
		                              &           90k-120k            &              128               &               32               &           Coarse+Accurate           &             0.06             &           16+1            &           4e-5            &           2e-5            &                                         &                  0.02                  \\ \hline
		    \multirow{2}{*}{BFM}      &             0-50k             &               32               &              128               &             Coarse+Fine             &             0.12             &           16+16           &           4e-4            &           4e-4            &          \multirow{2}{*}{0.5}           &                   0                    \\
		                              &           50k-110k            &               64               &              112               &           Coarse+Accurate           &             0.06             &           16+1            &           4e-4            &           4e-4            &                                         &                  0.05                  \\ \hline
	\end{tabular}}
\end{table*}

We leverage in PyTorch \cite{paszke2019pytorch} to implement the proposed method. All the experiments are trained for 60-80 hours on a server with four NVIDIA RTX3090 GPUs. We utilize the Adam \cite{kingma2014adam} optimizer to train the model. In Table \ref{tab:training details}, we list the specific experiment details of the different training stages. For CARLA and CelebA datasets, the training process is consisted of three phases, a $32 \times 32$ resolution phase, a $64\times 64$ resolution phase and a $128 \times 128$ resolution phase. In terms of the BFM dataset, training process only include the first two phases in the above datasets. Our sampling strategy of camera positions during training is adaptive to the datasets. For CARLA, we use the real distribution as described above. Since CelebA is a collection dataset of real camera data without camera parameters, it is impossible to obtain the real distribution of camera poses. For a fair comparison, we therefore use an estimated normal distribution used in PI-GAN, with a vertical standard deviation of $ 0.155 $ radians and a horizontal standard deviation of $ 0.3 $ radians. And for BFM, we use the same camera pose distribution as CelebA.

\subsection{Main Results and Qualitative Analysis}
\begin{figure*}[!h]
	\centering
	\includegraphics[width=\linewidth]{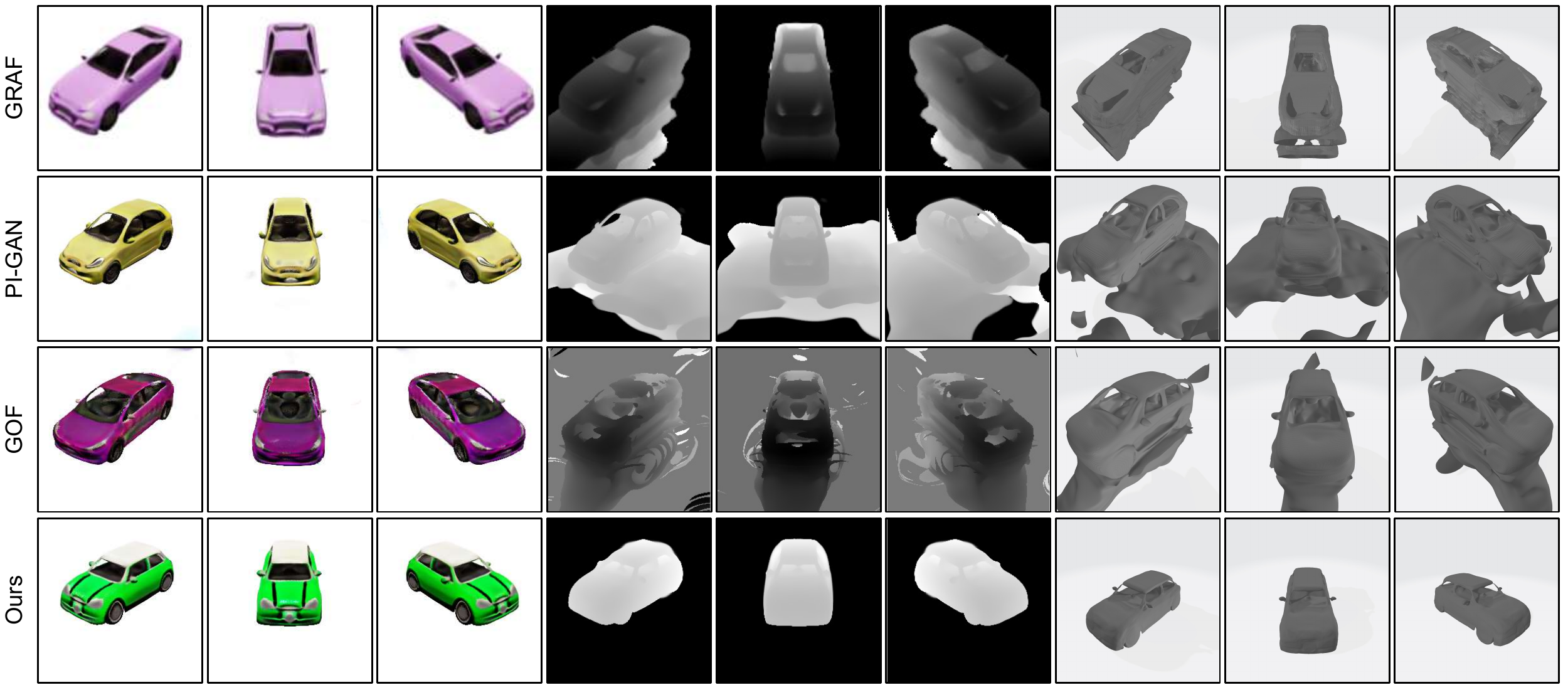}
	\caption{The comparison of CARLA dataset. The first three columns are generated RGB images from different observation directions. The next three columns are the corresponding depth map in the coordinate system. The last three columns are the corresponding 3D mesh.}
	\label{fig:carla_comparsion}
\end{figure*}

\begin{figure*}[!h]
	\centering
	\includegraphics[width=\linewidth]{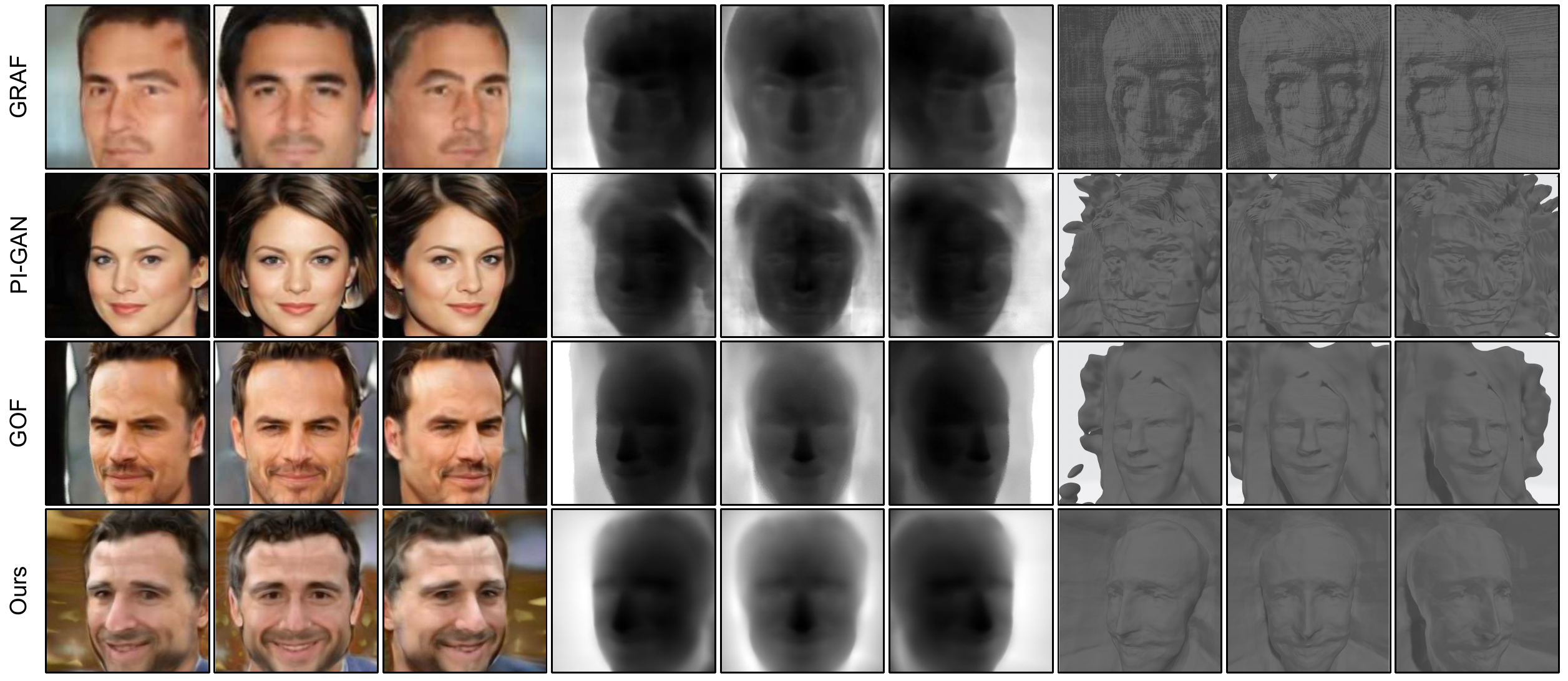}
	\caption{The comparison of CelebA dataset. The first three columns are generated RGB images from different observation directions. The next three columns are the corresponding depth map in the coordinate system. The last three columns are the corresponding 3D mesh.}
	\label{fig:celeba_comparsion}
\end{figure*}

\begin{figure*}[!h]
	\centering
	\includegraphics[width=\linewidth]{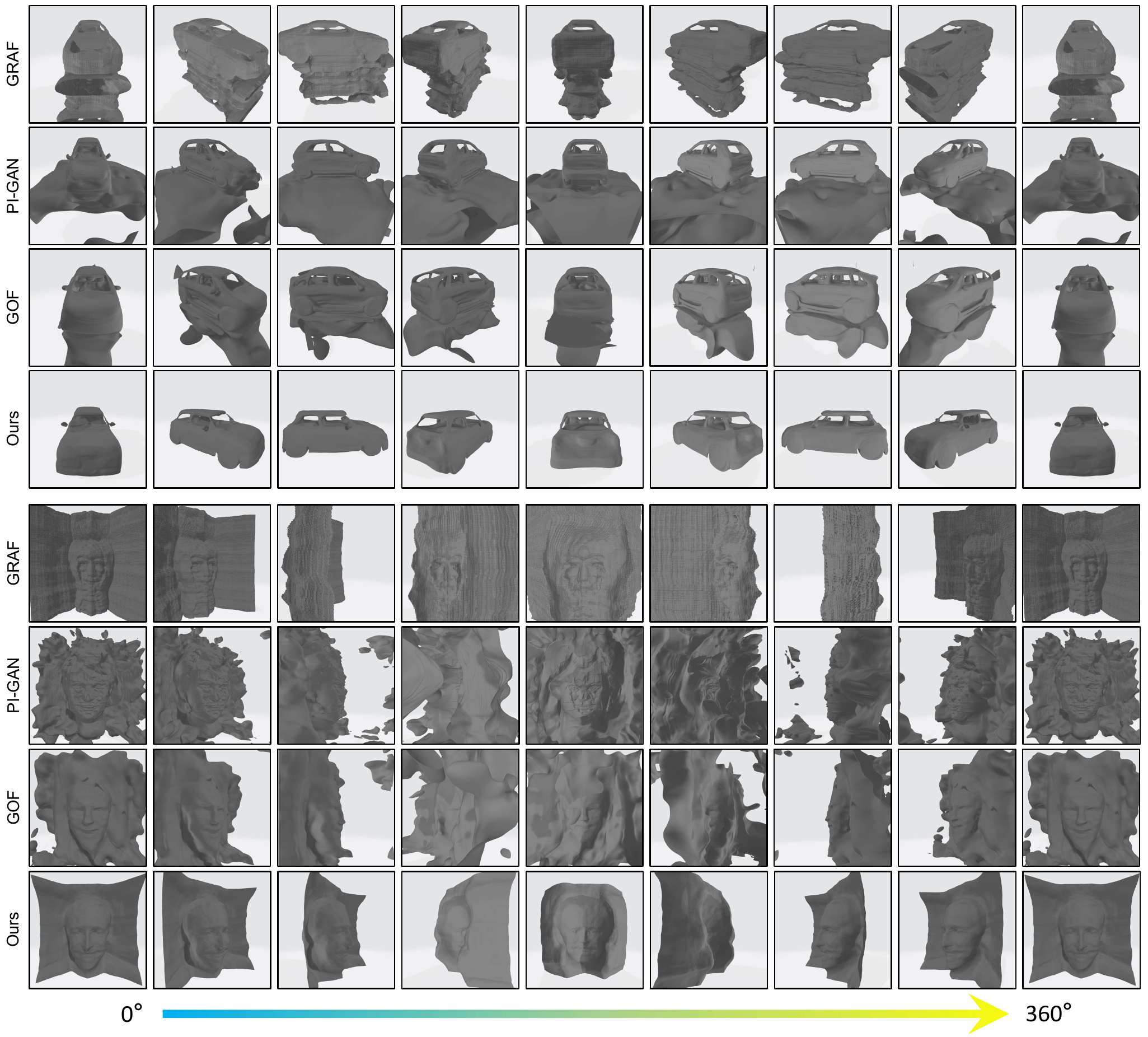}
	\caption{The rotation image of generated meshes on CARLA (top four rows) and CelebA (bottom four rows) datasets. And it is a full angle rotation from left to right.}
	\label{fig:mesh_rotation}
\end{figure*}

\begin{figure*}[!h]
	\centering
	\includegraphics[width=\linewidth]{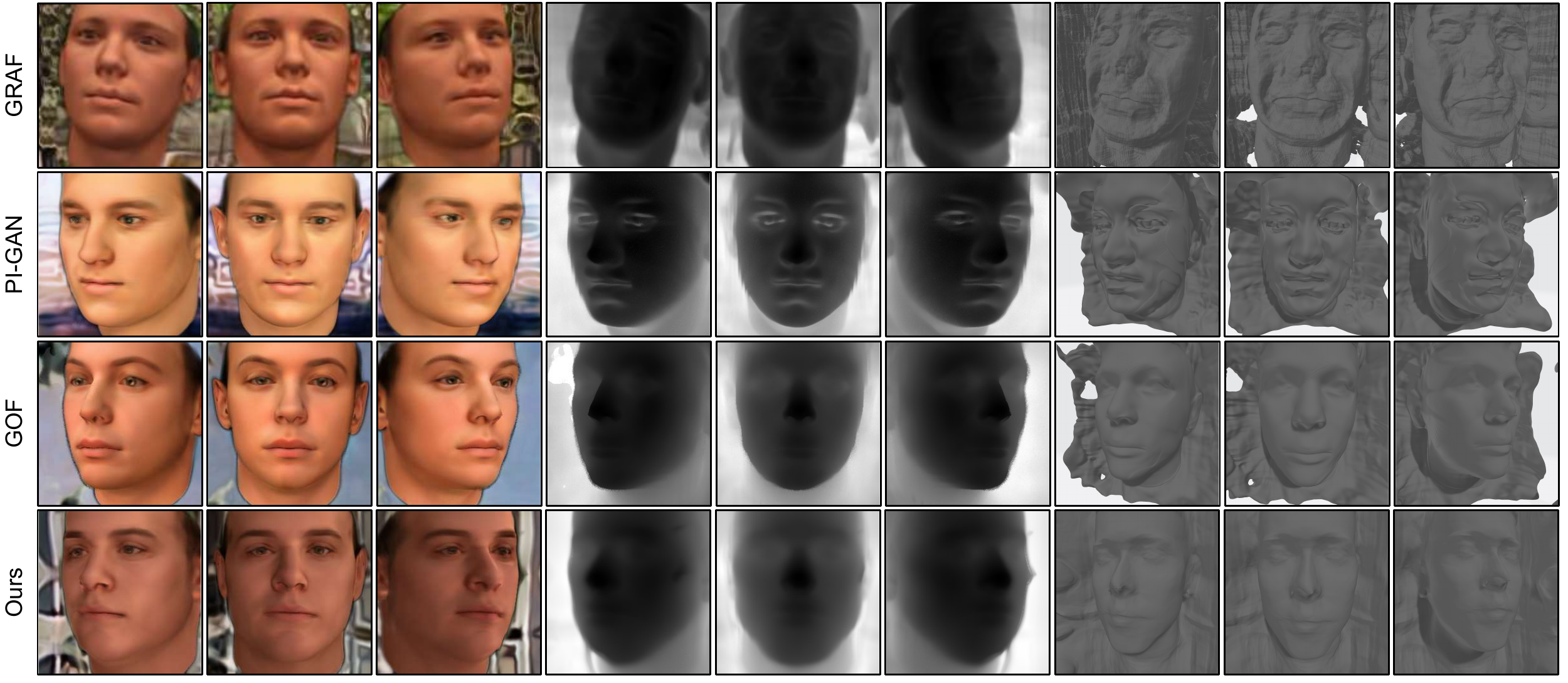}
	\caption{The comparison of BFM dataset. The first three columns are generated RGB images from different observation directions. The next three columns are the corresponding depth map in the coordinate system. The last three columns are the corresponding 3D mesh.}
	\label{fig:BFM_comparsion}
\end{figure*}

\begin{figure*}[!h]
	\centering
	\includegraphics[width=\linewidth]{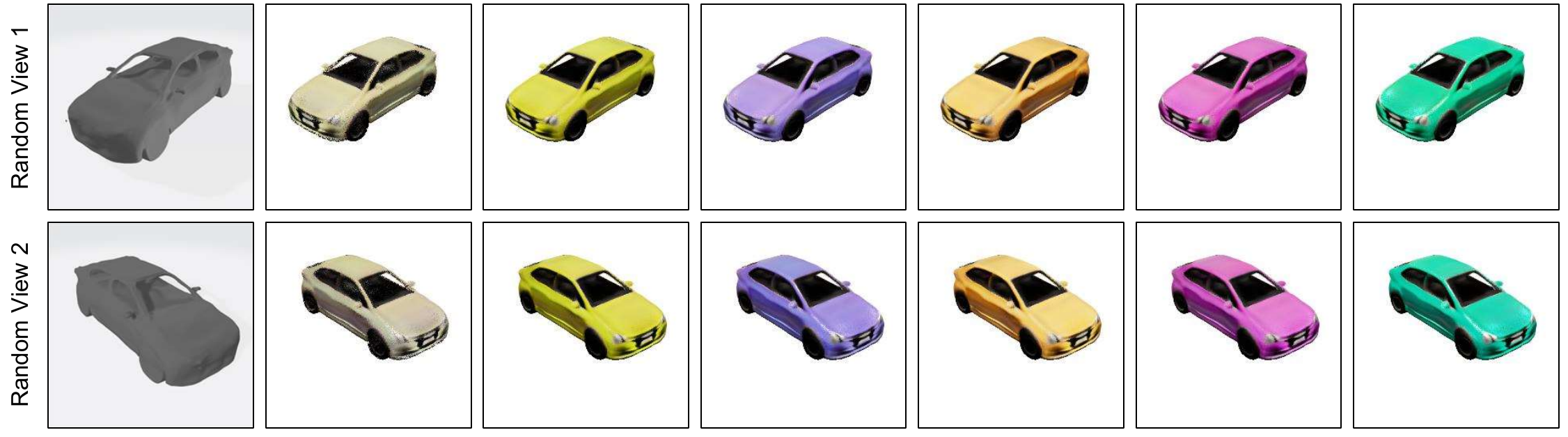}
	\caption{The color change of a given fixed shape. We first fix $\bm{z_s}$ to generate a 3D model. Then we use different $\bm{z_c}$ to render different surface color.}
	\label{fig:color_change}
\end{figure*}

We choose to use the visual results of the generated 3D model to finish the comparison of 3D object generation quality. It is worth noting that we have not carefully selected the generated 3D models. The 3D models generated by our method are all the smooth and realistic high quality models we present. And the 3D models generated by the other methods suffer from spatial clutter or unevenness of the model surface. As for the quality of 3D-Aware image synthesis, we use the Fréchet Inception Distance (FID) \cite{heusel2017gans} and Kernel Inception Distance (KID) \cite{binkowski2018demystifying} metrics, which are widely used in the field of image synthesis, to evaluate. The FID and KID measure the distance between the distribution of the generated data and the training data. And we use frames per second (FPS) to measure the speed of generating 3D-Aware images.

We compared our method with three mainstream approaches, including GRAF \cite{schwarz2020graf}, PI-GAN \cite{chan2021pi} and GOF \cite{xu2021generative}. In Fig. \ref{fig:carla_comparsion}, we show a visual comparison between our method and the state-of-the-art in terms of 2D image, depth map and 3D mesh on the CARLA dataset. As can be seen from Fig.~\ref{fig:carla_comparsion}, for the depth map and 3D mesh, there is much mess in the bottom of car yielded by these three methods. Besides, the bodywork of car is hardly generated smoothly and shows a severe ripple-like effect (zoom in for a better view). We believe it is because previous methods directly output the volume density or opacity of a point, making it difficult to obtain a effective supervision of chaotic area. By contrast, our method allows for the direct representation of a continuous surface of an object in space by the zero-level set of the SDF, which is equal to place a numerical constraint on the position away from the surface of the object and enhances the relationships between points and surface in space. Therefore, we avoid the mess around the car effectively. To further validate the effectiveness of the proposed method, in the top four rows of Fig.~\ref{fig:mesh_rotation}, we demonstrate the full angles rotation of one of the generated mesh models on the CARLA dataset. It can be seen that our method shows obvious advantages over existing methods in generation quality. 

In Fig.~\ref{fig:celeba_comparsion}, we visually compare our method with existing cutting-edge methods on the CelebA dataset. One can see that although the previous methods can generate continuous 3D-aware images, the corresponding spatial accuracy is inferior to our method. In terms of mesh performance GRAF exported, there are many small lumpy patches and incorrect depressions, resulting in undesirable quality. Although the mesh results yielded by PI-GAN are significantly improved, there are still many abnormal stripes. GOF proposes the Opacity Regularization Loss, which encourages all output of neural network to be zero or one. It is advisable to forces only points very close to the surface to be with non-zero weights when calculating the color. However, such a constraint makes network output near the surface vary drastically with the micro-variation of the input, posing more challenges on fitting the relationships in the space.  Unlike GOF, our proposed method fits the SDF of an object, which enables output of neural network to vary in a smooth form as the input changes, while preserving the principal advantage of GOF. As evidenced in the bottom four rows of Fig.~\ref{fig:mesh_rotation}, we demonstrate the full angles rotation of the generated mesh model on the CelebA dataset. We can see that the mesh generated by our model is more realistic and natural, while enjoying smooth and spotless merits. The other methods, however, are challenged by clutters spread the space due to the lack of explicit constraints in space. Fig. \ref{fig:BFM_comparsion} presents a similar visual trend on BFM dataset.

\begin{table*}[t!]
	\centering
	\caption{Comparison with other baseline's on FID, KID and FPS at $128 \times 128$ resolution.}
	\label{tab:Overall result}
	\resizebox{\linewidth}{!}{
		\begin{tabular}{c|ccc|ccc|ccc}
			\hline
			  \multirow{2}{*}{Method}   &                        \multicolumn{3}{c|}{CARLA}                         &                        \multicolumn{3}{c|}{CelebA}                        &                         \multicolumn{3}{c}{BFM}                          \\ \cline{2-10}
			                            & FID $\downarrow$ & KID $\downarrow$ & \multicolumn{1}{l|}{FPS $\uparrow$} & FID $\downarrow$ & KID $\downarrow$ & \multicolumn{1}{l|}{FPS $\uparrow$} & FID $\downarrow$ & KID $\downarrow$ & \multicolumn{1}{l}{FPS $\uparrow$} \\ \hline
			GRAF \cite{schwarz2020graf} &      39.71       &       1.94       &                4.76                 &      32.28       &       1.99       &                4.75                 &      31.10       &       2.04       &                4.74                \\
			 PI-GAN \cite{chan2021pi}   &      26.73       &       0.88       &                3.48                 &      15.04       &       0.41       &                13.00                &      15.93       &       0.49       &               12.98                \\
			GOF \cite{xu2021generative} &      26.69       &       0.89       &                5.82                 &      15.58       &       0.44       &                8.02                 &      14.28       &       0.42       &                7.74                \\
			         SDF-3DGAN          &  \textbf{24.93}  &  \textbf{0.75}   &           \textbf{13.31}            &  \textbf{14.53}  &  \textbf{0.35}   &           \textbf{15.08}            &  \textbf{12.01}  &  \textbf{0.34}   &           \textbf{15.00}           \\ \hline
		\end{tabular}}
\end{table*}

Table \ref{tab:Overall result} summarizes quantitative comparisons between the proposed method and current SOTA methods in terms of Fréchet Inception Distance (FID) \cite{heusel2017gans} and Kernel Inception Distance (KID) \cite{binkowski2018demystifying} on the datasets (CARLA, CelebA and BFM). For a fair comparison, we generate images at $128 \times 128$ resolution to calculate the metrics of FID and KID. Without bells and whistles, the quantitative performance shows that our method significantly performs better than the previous method. And performance in terms of frames per second (FPS) further proves the computation efficiency of our proposed new rendering pipeline.

In Fig. \ref{fig:color_change}, we also show the effect of changing $\bm{z_c}$ for a given fixed $\bm{z_s}$, \ie, the 3D model is the same, but the appearance color of the model is constantly changed.

\subsection{Ablation Studies}

\subsubsection{\textbf{Accurate Sampling}}
\label{section:Accurate sampling}

\begin{figure*}[!h]
	\centering
	\includegraphics[width=7in]{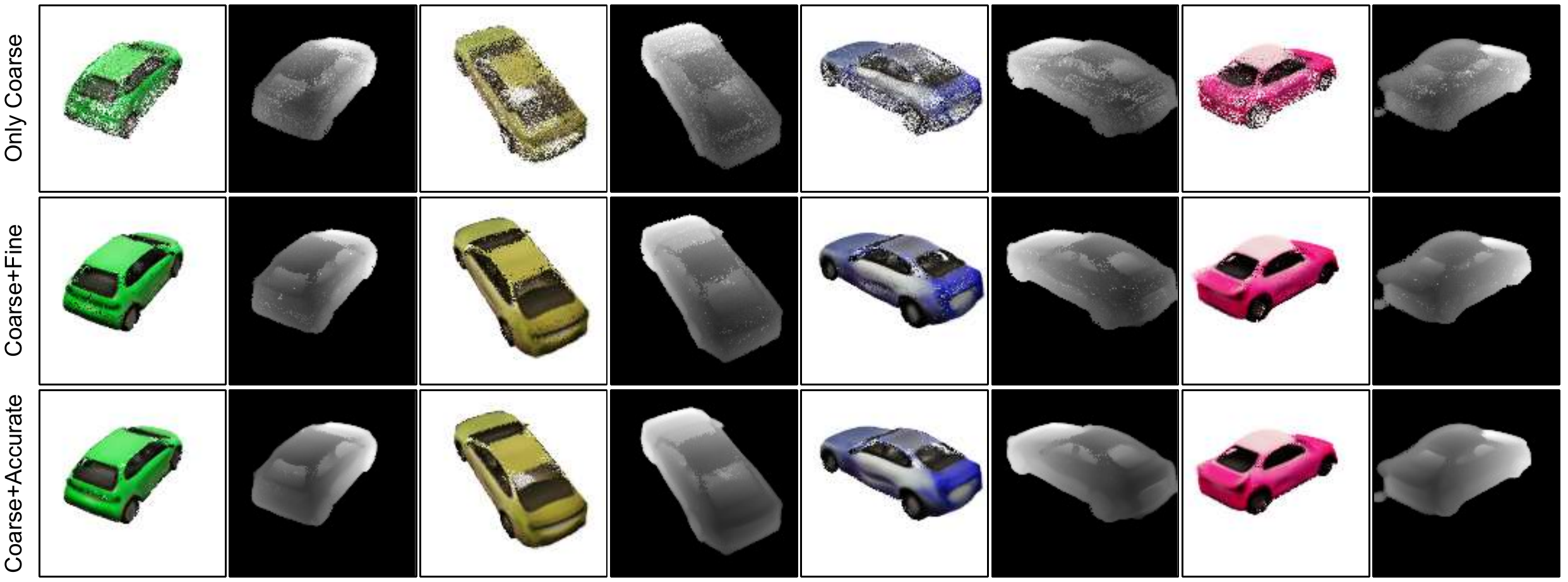}
	\caption{The comparison among only coarse sampling, coarse+fine sampling and coarse+accurate sampling.}
	\label{fig:ablation_accuracy_sampling}
\end{figure*}

To verify the effect of our proposed sampling strategy,  we perform experiments to make comparisons with the sampling method adopted by the original NeRF and PI-GAN. We choose a model, which is trained for the first stage on CARLA dataset at $ 32 \times 32 $ resolution using coarse-to-fine sampling, as the baseline. The explanation for this is that a rudiment of SDF is just responsible for shape. And we observe that as the $\beta$ in Equ.~(\ref{eq:map_function}) rises, the problems depicted in Fig. \ref{fig:two_error} start to emerge. In Fig. \ref{fig:ablation_accuracy_sampling}, we show the comparisons of ``Only Coarse'' sampling, ``Coarse+Fine'' sampling and our ``Coarse+Accurate'' sampling. As can be seen, the manners using coarse sampling and two rounds of ``Coarse+Fine'' sampling (first two lines in Fig. \ref{fig:ablation_accuracy_sampling}) show numerous small dots spread the surface, resulting in the discontinuous surface of an object. Differently, our method (last line in Fig. \ref{fig:ablation_accuracy_sampling}) not only solves the sampling ambiguity problem (shown in Fig. \ref{fig:two_error}), but also significantly improves the sampling efficiency with less computation. Assuming that coarse sampling requires $N$ points and fine sampling requires $M$ points, our method succeeds to reduce the sampling number from $N+M$ to $N+1$, which is ascribed to the fact that SDF serves as intermediate representations in 3D space and significantly reduce computational burden.

\subsubsection{\textbf{Speed Comparison of Different Renderer}}

\begin{table}[]
	\centering
	\caption{Comparison of different rendering methods.}
	\label{tab:render method comparison}
	\begin{tabular}{c|c}
		\hline
		       Method        & FPS $\uparrow$ \\ \hline
		Conventional  Method &      2.65      \\
		     Our Method      &     13.31      \\ \hline
	\end{tabular}
\end{table}

To verify the effectiveness of our proposed rendering pipeline, we use an already trained SDF generator to compare the speed of rendering a $128 \times 128$ image when two methods achieving similar generated image quality. Therein, one is the volume rendering method that most of the traditional methods used. And another is our method, which is optimized by the SDF mathematical property. In Table \ref{tab:render method comparison}, the results show that our designed rendering method can significantly reduce the required rendering time, which in turn saves training time. This improvement is due to the fact that our renderer can use fewer points to locate the surface of objects more accurately, while the previous methods must use the enough points.

\subsubsection{\textbf{Eikonal Loss Weight}}
Eikonal Loss is the key to ensuring that our network can be seen as an implicit SDF. It is obviously that the gradient of the SDF function with respect to the input coordinates should be $1$ everywhere. Therefore, if the Eikonal Loss is too large, the network cannot be seen as an implicit SDF. We conduct some experiments on CARLA dataset with different $\lambda_{Eikonal}$. In Table \ref{tab:Eikonal Loss Weight}, too small the $\lambda_{Eikonal}$ will result in too large the Eikonal Loss, which prevents the neural network from acting as an implicit SDF. When $\lambda_{Eikonal}$ is too large, it will make the network pay too little attention to the quality of the generated image, leading to a decrease in the quality of the generated 3D-aware image. Therefore, we choose $0.5$ as the weight for Eikonal Loss.

\begin{table}[]
	\centering
	\caption{Comparison of different Eikonal Loss Weights.}
	\label{tab:Eikonal Loss Weight}
	\begin{tabular}{c|cc}
		\hline
		$\lambda_{Eikonal}$ & FID $\downarrow$ & Eikonal Loss $\downarrow$ \\ \hline
		        0.1         &        -         &           1.02            \\
		        0.5         &      24.93       &           0.04            \\
		        1.0         &      26.32       &           0.01            \\
		        2.0         &      30.03       &           0.01            \\ \hline
	\end{tabular}
\end{table}

\subsubsection{\textbf{Normal Loss Weight}}
\begin{figure}[!h]
	\centering
	\includegraphics[width=\linewidth]{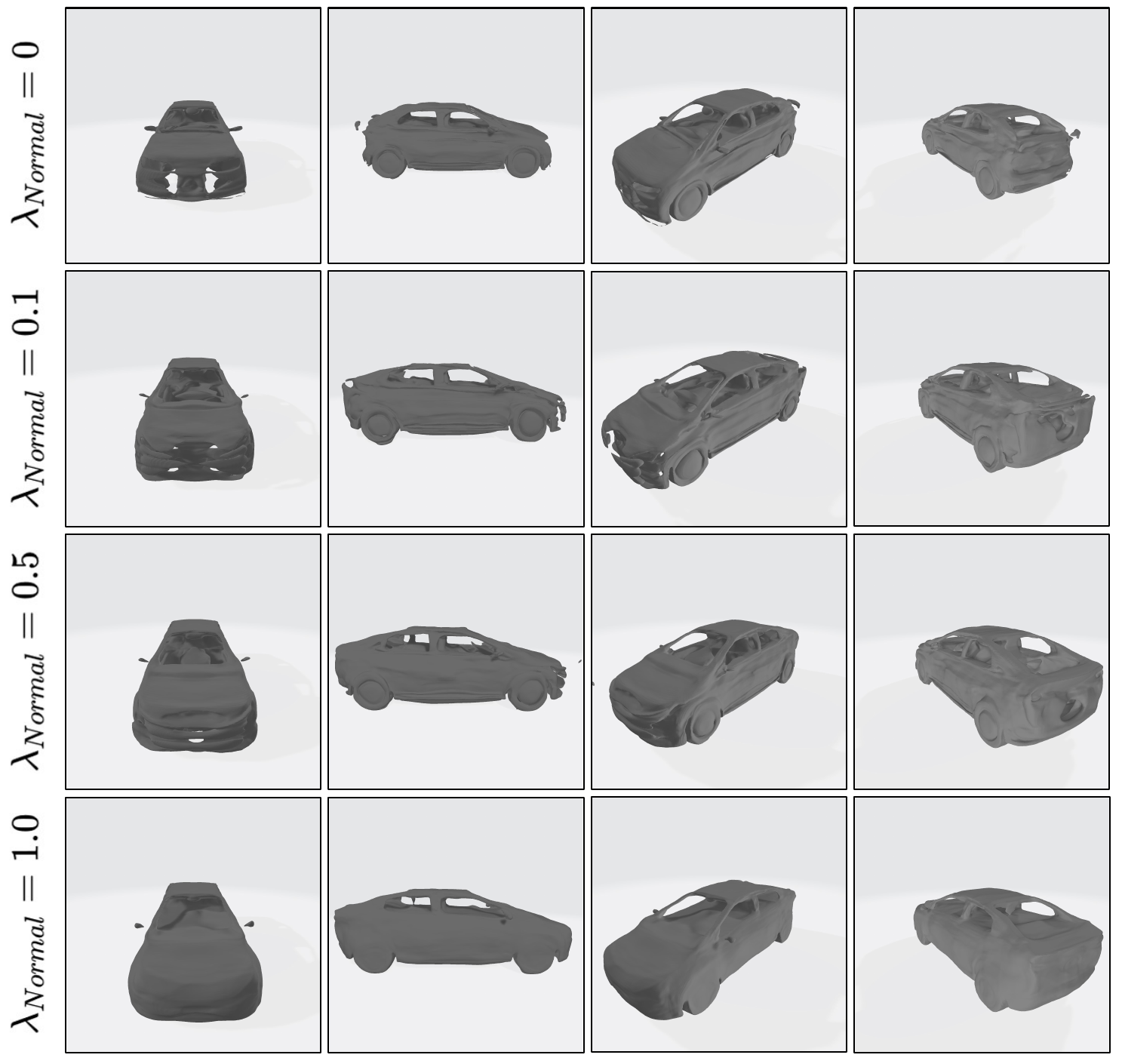}
	\caption{The comparison of different normal coefficients.}
	\label{fig:ablation_normal}
\end{figure}

As we described earlier, we can manipulate the smoothness of the generated 3D model by $ \ell_2 $ Normal Loss, thanks to the fact that our model explicitly perceives surface of the generated 3D model. Different datasets require different coefficients, but the general trend is the same. We have therefore chosen the CARLA dataset as an example. Fig. \ref{fig:ablation_normal} illustrates the smoothness of SDF after conversion to mesh conditioned on different Normal Loss coefficients. It is quite obvious that the larger factor is, the smoother the surface is. And when this loss item is removed,  we can see that the surface of the model degrades dramatically. In the Table \ref{tab:Normal Loss Weight}, we find the generation quality of 3D-Aware images decreases as $\lambda_{Normal}$ rises. Considering the visual results are good enough with a setting of 1.0, so we set it to 1.0 on balance.

\begin{table}[]
	\centering
	\caption{Comparison of different Normal Loss Weights.}
	\label{tab:Normal Loss Weight}
	\begin{tabular}{c|c}
		\hline
		$\lambda_{Normal}$ & FID $\downarrow$ \\ \hline
		       0.0         &      24.63       \\
		       0.1         &      24.82       \\
		       0.5         &      24.87       \\
		       1.0         &      24.93       \\ \hline
	\end{tabular}
\end{table}

\section{Conclusion}\label{sec:conclusion}
To improve 3D object generation task, we deliver a new method termed SDF-3DGAN, which introduce implicit Signed Distance Function (SDF) as the representation method of 3D object in the generative field. We apply SDF for higher quality representation of a 3D object in space and design a new SDF neural renderer. In the training process, we first generate the objects represented by implicit SDF. Then, we use our new rendering pipeline to render them to 2D images at arbitrary angles to apply GAN training method with 2D images in the dataset. We utilize the mathematical property of SDF solves the problems of sampling ambiguity when the number of sampling points is too small. Therefore, we can use the less points to finish higher quality rendering, which saves our training time. At the same time, our renderer can locate the surface point easily, which make us can apply normal loss to control the normal vector variation to make our generated object surface smoother and more realistic. Finally, quantitative and qualitative experiments conducted on public benchmarks demonstrate favorable performance against the state-of-the-art methods in 3D object generation task and 3D-Aware image synthesis task.

The limitation of our method is that in training stage, the resolution of rendered 3D-aware images is still not high enough. This will affect the generation quality of more surface details. Therefore, in the future research, we will shift focus on exploring the implicit representation generator on the higher resolution.

\bibliographystyle{IEEEtran}
\bibliography{ref}

\newpage

\vfill

\end{document}